\documentclass[10pt,twocolumn,letterpaper]{article}

\usepackage[pagenumbers]{wacv} 

\usepackage{amsmath}
\usepackage{amssymb}
\usepackage[utf8]{inputenc} 
\usepackage[T1]{fontenc}    
\usepackage{url}            
\usepackage{booktabs}       
\usepackage{amsfonts}       
\usepackage{nicefrac}       
\usepackage{microtype}      
\usepackage{xcolor}         
\usepackage[inline]{enumitem}
\usepackage{graphicx}
\usepackage{xspace}
\usepackage{subcaption}
\usepackage{float}
\usepackage{wrapfig}
\usepackage{array}
\usepackage{placeins}
\usepackage{tikz}
\usepackage{multirow}
\usepackage{makecell}

\newcommand{\best}[1]{\textbf{#1}}
\newcommand{\second}[1]{\underline{#1}}

\newcommand{\condenseparagraph}[1]{{\vspace*{2mm}\noindent\textbf{#1}\quad}}

\definecolor{darkmagenta}{RGB}{186, 0, 186}
\definecolor{slateblue}{RGB}{90, 70, 230}

\usepackage[pagebackref,breaklinks,colorlinks]{hyperref}

\usepackage[capitalize]{cleveref}
\crefname{section}{Sec.}{Secs.}
\Crefname{section}{Section}{Sections}
\Crefname{table}{Table}{Tables}
\crefname{table}{Tab.}{Tabs.}

\definecolor{greencode}{RGB}{13, 144, 79}



\newcommand{\templatecaption}{T^{\rm class}}
\newcommand{\imagecaption}{{T^{\rm image}}}
\newcommand{\batchcaption}{{T^{\rm group}}}
\newcommand{\image}{x}
\def\modelname{\textsc{LatteCLIP}\xspace}

\def\wacvPaperID{594} 
\def\confName{WACV}
\def\confYear{2025}

\begin{document}

\title{\modelname: Unsupervised CLIP Fine-Tuning via LMM-Synthetic Texts}

\author{
	Anh-Quan Cao$^{2}\thanks{The main work was done while interning at Amazon.}$ \quad
	  Maximilian Jaritz$^1$ \quad
       Matthieu Guillaumin$^1$ \quad Raoul de Charette$^2$ \quad Loris Bazzani$^1$ \vspace{0.2cm}\\
	$^1$Amazon \qquad $^2$Inria \\
}

\maketitle

\begin{abstract}
   Large-scale vision-language pre-trained (VLP) models (\eg, CLIP~\cite{CLIP}) are renowned for their versatility, as they can be applied to diverse applications in a zero-shot setup. However, when these models are used in specific domains, their performance often falls short due to domain gaps or the under-representation of these domains in the training data. While fine-tuning VLP models on custom datasets with human-annotated labels can address this issue, annotating even a small-scale dataset (\eg, 100k samples) can be an expensive endeavor, often requiring expert annotators if the task is complex. {To address these challenges, we propose \modelname, an unsupervised method for fine-tuning CLIP models on classification with known class names in custom domains, without relying on human annotations.} 
   Our method leverages Large Multimodal Models (LMMs) to generate expressive textual descriptions for both individual images and groups of images. These provide additional contextual information to guide the fine-tuning process in the custom domains. Since LMM-generated descriptions are prone to hallucination or missing details, we introduce a novel strategy to distill only the useful information and stabilise the training. {Specifically, we learn rich per-class prototype representations from noisy generated texts and dual pseudo-labels}. Our experiments on 10 domain-specific datasets show that \modelname outperforms pre-trained zero-shot methods by an average improvement of +4.74 points in top-1 accuracy and other state-of-the-art unsupervised methods by +3.45 points. 
\end{abstract}

\section{Introduction}
\label{sec:intro}

\begin{figure}[ht]
    \centering
    \includegraphics[width=\linewidth]{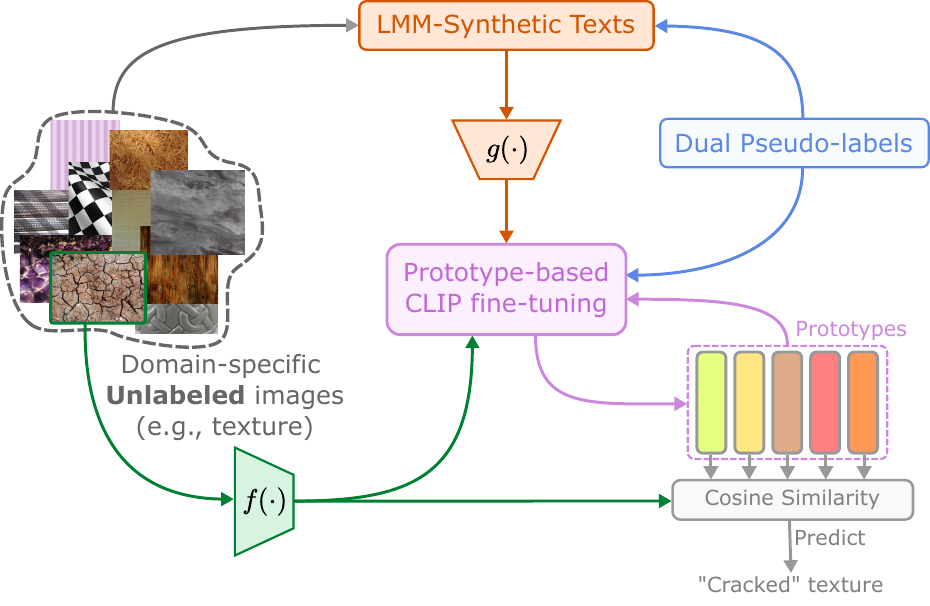}
    \vspace{-0.6em}
    \caption{\textbf{Overview of \modelname.}~Our prototype-based method leverages different types of pseudo-labels and LMM-synthetic texts for improved unsupervised CLIP fine-tuning on domain-specific datasets (\textit{e.g.,} texture). During inference, image features are compared with prototypes to generate predictions. Here, $f(\cdot)$ and $g(\cdot)$ are the CLIP image and text encoders, respectively.}
    \label{fig:teaser}
\end{figure}

Large-scale vision-language pre-training~\cite{CLIP} has emerged recently and demonstrated impressive generalization performance on various downstream tasks~\cite{podo, jia2021scaling, li2021align, simple, patashnik2021styleclip}, especially in zero-shot classification~\cite{CLIP, waffleclip}. This success is attributed to its robust visio-linguistic representation, learned from a vast amount of large-scale web-scraped datasets~\cite{laion5b}. 
However, these models often face challenges in specialized domains due to domain discrepancies and insufficient representation in the training data. Prior studies have demonstrated {improvements on custom datasets through supervised fine-tuning~\cite{wiseft,FLYP} or few-shot learning}~\cite{zhou2022learning,silva2024closer}.
Nevertheless, acquiring human-annotated labels is costly, even for {relatively} small datasets (\textit{e.g.,} ~100k samples), and often requires expert annotators {for complex tasks}. 
{To address this, we propose \modelname, which fine-tunes CLIP for classification on unlabeled training data to maximize performance on a test set from the same domain. Here, a \emph{domain} refers to a set of shared characteristics within a dataset (e.g., cars, flowers, textures). Like in Unsupervised Domain Adaptation (UDA)~\cite{ganin2016domain, wei2021metaalign, hoyer2023mic}, we consider the list of class names to be known a priori.} An overview of \modelname is shown in~\cref{fig:teaser}.

{
Recent progress of Large Language Models~(LLMs)~\cite{openai2024gpt4, anthropic2024claude, geminiteam2024gemini, jiang2024mixtral, touvron2023llama} and Large Multimodal Models~(LMMs)~\cite{bai2023qwen, LLAVA} have led to a fundamental shift in training and fine-tuning methodologies. 
The research community is transitioning from a class-focused paradigm towards a more descriptive approach, where data is annotated with detailed textual descriptions for training, and rich textual answers are provided at inference time.
Consequently, an increasing number of methods~\cite{laclip, lai2024veclip, synthclip} now leverage synthetically-generated text from LMMs as an additional source of supervision or contextual information to improve performance. 
Similar to these approaches, we harness the power of LMMs to generate descriptions for training, but with a strong emphasis on producing more expressive descriptions. Instead of only generating per-image descriptions, we also generate descriptions for groups of images, capturing their common characteristics, as well as class-level descriptions for all images within a category. These descriptions provide better contextual information, offering richer supervision for training, leading to improved classification accuracy in specific domains compared to the limited information from pseudo-labels and label propagation~\cite{pseudolabel,hu2024reclip}.}

{However, directly fine-tuning CLIP with LMM-generated texts leads to poor performance due to CLIP overfitting to hallucinations and noise present in these texts. To address this, we propose a fine-tuning framework based on prototype learning~\cite{aamodt1994case, newell1972human}, where classes are represented as a set of prototypes, typically as feature vectors.
Prototypes provide better control and interpretability of class representations through direct manipulation in the embedding space, helping regulate the influence of each synthetic description during training.
}
{To further improve the per-class prototype representations, we combine the synthetic texts with two types of pseudo-labels derived from both zero-shot and fine-tuning models. The zero-shot model offers better generalization thanks to pre-training knowledge, while the fine-tuning model provides stronger in-domain performance.
During inference, these prototypes are compared with image features for classification. As LMMs are only employed during during fine-tuning, the inference time remains consistent with standard CLIP methods. We validate the effectiveness of our method across 10 domain-specific datasets. Compared to pre-trained CLIP models, \modelname achieves an average improvement of +4.74 points in top-1 accuracy, surpassing other unsupervised fine-tuning baselines by +3.45 points.}

Our contributions can be summarized as follows:
\begin{itemize}[noitemsep,topsep=0pt]
    \item We propose \modelname, a novel method that synthesizes multiple types of image descriptions to enhance the unsupervised fine-tuning of CLIP models on domain-specific datasets, leveraging the language expressiveness of  LMMs.
    \item To make training robust to noisy texts and pseudo-labels, we employ a prototype framework with a momentum update, enabling us to control the influence of synthetic text features. To further refine the useful image descriptions, we introduce a Dynamic Feature Mixer module that assigns higher weight to important text, resulting in better-combined text embeddings.
    \item We show that mixing pseudo-labels from zero-shot model and fine-tuning model significantly improves performance; the former preserves pre-trained knowledge, while the latter improves the accuracy on the target distribution. Experiments show that \modelname significantly outperforms all baselines on average across 10 domain-specific datasets.
\end{itemize}

\section{Related works}

\condenseparagraph{Adapting CLIP for Classification.}
CLIP-based methods~\cite{CLIP,yao2021filip,yu2022coca,zhai2023sigmoid} exhibit competitive zero-shot classification performance. For further improvement on downstream classification datasets, CLIP can be adapted to close the gap between pre-trained representations and specific domains.

In few-shot learning, one has access to a small number of labels, typically between 1-16 samples per class, and many works have adapted it to CLIP~\cite{gao2024clip,zhang2022tip,zhou2022learning,zhu2023not,qian2024intra,silva2024closer,martin2024transductive,gondal2024domain,zanella2024low, fahes2024prolip}. Prototypical learning~\cite{snell2017prototypical} is a seminal work in few-shot learning and builds an average embedding (prototype) for each class. During inference, one then matches the test sample to the nearest prototype. This concept recently re-emerged for adapting CLIP, by building a cache model holding the knowledge from the few-shot training set~\cite{zhang2022tip,zhu2023not}. Different from that, we continuously update the prototypes with momentum during the training process with multimodal features from \textit{unsupervised} texts and images. Other works leverage prompt learning~\cite{zhou2022learning} or efficient fine-tuning~\cite{zanella2024low, fahes2024prolip}.

Supervised fine-tuning methods require a significant amount of labeled examples for training~\cite{CLIP,FLYP,wiseft,wei2023improving,lpft}. Linear probing~\cite{CLIP,wiseft} is a simple technique that trains a classifier on top of frozen image features, but can lead to worse results due to overfitting. This problem has been tackled by using two-step training schedules of linear probing and full fine-tuning~\cite{lpft}, masked image modeling~\cite{wei2023improving} and by fine-tuning with contrastive loss by aligning the image with a template text including the class label (FLYP~\cite{FLYP}). In contrast to FLYP, we add LMM-generated descriptions to the contrastive loss, and stabilize unsupervised training by learning prototypes with momentum.

Different from few-shot and fine-tuning, we focus on the challenging scenario of unsupervised fine-tuning, where no labels are available, because they are too costly to annotate.

\condenseparagraph{Unsupervised Model Adaptation}
Our Unsupervised fine-tuning task is related to Unsupervised Domain Adaptation, where one typically reduces the discrepancy between the source and target data. However, lately, the task of source-free domain adaptation (SFDA) has emerged, where target adaptation is performed without access to the source data, see survey paper~\cite{li2024comprehensive}. Many methods exploit that the source model can partially generalize to the target domain, and fine-tune with pseudo-labels~\cite{liang2020we}, adversarial learning~\cite{li2020model}, historical contrastive learning~\cite{huang2021model} or mixup~\cite{kundu2022balancing}. While \cite{huang2021model} perform momentum contrastive learning~\cite{he2020momentum} on different image augmentations, we contrast image-text pairs. The above SFDA works train on a narrow source distribution. Instead, ReCLIP~\cite{hu2024reclip} leverages CLIP, which is pre-trained on wide-distribution large-scale data. ReCLIP leverages pseudo labels, cross entropy between separate modalities and focus on transductive setup (train/test on test set). 
{Test-time adaptation methods~\cite{shu2022test, rlcf, sun2020test, liu2021ttt++, wang2020tent} update the model to align with the target distribution at test time using a single image in self-supervised manner, requiring optimization at inference. Unlike these approaches, we leverage LMM-generated texts to maximize test performance by fine-tuning model parameters on unlabeled training data, thus keeping the model parameters fixed during testing.}

\condenseparagraph{LMMs for Synthetic Labels.}
Using synthetically-generated labels and textual descriptions is becoming a standard in the field, because of the general availability of LLMs and LMMs that can be prompted and guided with task-specific examples~\cite{ouyang2022training,tan2024large,LLAVA,lai2024veclip,synthclip,laclip,achiam2023gpt}.
This provides an opportunity for VLP, that typically uses large-scale image-text pair data scraped from the internet, \eg, LAION-5B~\cite{laion5b}. 
Instead of using noisy and inconsistent captions or annotating a large set, we synthetically-generate descriptions.
While LaCLIP~\cite{laclip} rewrites existing captions with LLMs (text-only input), VeCLIP~\cite{lai2024veclip} prompts an LMM to caption the image, followed by LLM text processing. SynthCLIP~\cite{synthclip} synthesizes first the text, and then the images with text-to-image generative models. 
In our work, we leverage LMMs to caption images, focusing on fine-tuning rather than pre-training, combining the pre-trained model's knowledge with new synthetic captions in a balanced approach.

\section{Method}

Fine-tuning CLIP with combination of predefined templates, such as {\small \texttt{\color{greencode}``a photo of a [class].''}}, was shown to yield effective results when using ground-truth class labels~\cite{wiseft, lpft,FLYP}.
However, in the absence of ground-truth class labels, fine-tuning CLIP models with pseudo-labels~\cite{pseudolabel}, using FLYP~\cite{FLYP}, leads to limited improvements\footnote{\cref{tab:main-results} reports the performance of ``FLYP + pseudo labels'' where we show limited improvements with respect to CLIP across 10 datasets on average. We also observe performance drops on some datasets (\textit{e.g.,} Food101, Flower102).}
This can be caused by two factors.
First, the text employed as supervision, resulting from the combination of the template and pseudo-label, lacks expressivity and discriminativity. This is typically the case for classes that are not visually descriptive, such as types of land use (\eg, annual crop, industrial, \etc) or names of textures (\eg, paisley, sprinkled, \etc). 
Second, pseudo-labels are inherently noisy, which negatively affects the downstream classification performance due to domain shifts relative to the original training data. 

Our method, \modelname, addresses these limitations by proposing an expressive unsupervised text generation~(\cref{sec:text_gen}) and a prototype-based learning mechanism~(\cref{sec:proto-learning}) to mitigate noisy pseudo labels. 
{
To improve expressivity beyond pseudo-labeling, we build upon a recent LMM~\cite{LLAVA}, generating descriptions at multiple levels of contextual granularity, describing the individual image, group of similar images, and entire class. Individual image descriptions offer detailed though possibly extraneous information, which is addressed by group descriptions that capture shared characteristics of similar images, albeit with some noise. This noise is mitigated by class descriptions, which provide stable representations to address inconsistencies.
}
Equipped with such textual description, we additionally introduce a prototype-based learning framework that learns a set of class prototypes from the generated text features. These prototypes are updated in a momentum setting to produce a smooth optimization over the whole training set, reducing the effect of noise from outlier samples and incorrect synthesized texts.

\subsection{Expressive Text Generation with LMMs}
\label{sec:text_gen}

\begin{figure}
    \centering
    \includegraphics[width=\linewidth]{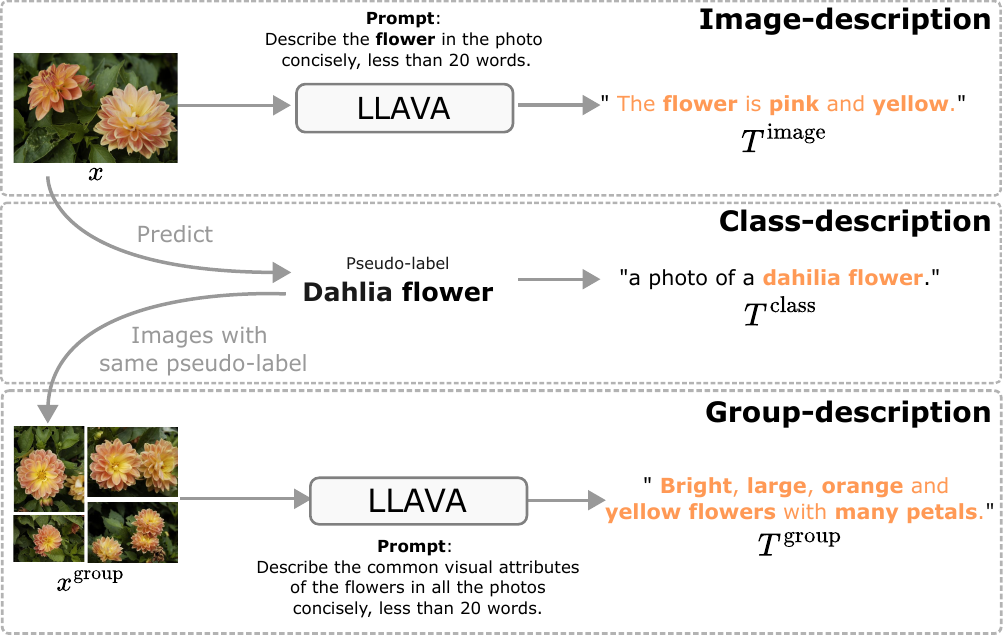}
    \vspace{-0.6em}
    \caption{\textbf{Text Generation with LMM.} In addition to the usual \textit{class-description} (middle), combining template text and pseudo-label, we leverage LMM~\cite{LLAVA} to generate \textit{image-description} (top) which provide more expressive visual description of the image. Further, by considering random group of images with the same pseudo-labels, we prompt~\cite{LLAVA} to capture shared characteristics as \textit{group-description} (bottom).
    }
    \label{fig:text_gen}
\end{figure}
Without access to ground-truth labels for training CLIP models, we must rely on noisy pseudo-labels.
Furthermore, class names alone often lack visual descriptiveness. Consequently, using only cross-entropy loss or solely relying on class names leads to suboptimal performance in our setting.
To address this challenge, we introduce a novel approach that leverages generated text to provide additional contextual information. 
In addition to the more standard \emph{class-description} mentioned above, we propose two additional ways using a recent LMM~\cite{LLAVA} to generate textual descriptions of images: \emph{image-description} and \emph{group-description}, as depicted in~\cref{fig:text_gen}.
These generated texts hold complementary information with increasing semantic abstraction, from class\footnote{Here, "class" refers to the class-description.}, to single image, to group of images, all of which help the model to learn more precise classification boundaries.
The \emph{image-description} texts provide detailed descriptions of individual images, capturing their unique characteristics and subtle features. The \emph{group-description} texts offer a comprehensive description representing the entire class, covering shared features and common attributes. 

Importantly, we found that the above mentioned descriptions are complementary to the use of template text with pseudo-label class, which we refer as \textit{class-description}. 
In fact, we later show that preserving this \textit{class-description} in the training process is crucial as our generated texts can be noisy due to missing details or hallucination. 
The combination of class-/image-/group-description provides a stable and reliable representation corresponding to the classes. 

More formally, for each image $\image$ we generate three texts, illustrated in \cref{fig:text_gen}, and defined as follows:\\
\noindent\textbf{Class-description} ($\templatecaption$) provides a consistent class representation using template
{\small \texttt{\color{greencode}``a photo of a [class].''}} where {\small \texttt{\color{greencode}[class]}} is substituted with the image pseudo-label $c$ obtained from a CLIP zero-shot.\\
\noindent\textbf{Image-description} ($\imagecaption$) captures unique features of image $\image$. We generate $\imagecaption$ by prompting LLAVA~\cite{LLAVA} with: 
{\small \texttt{\color{greencode}``Describe the [domain] in the photo concisely, using less than 20 words."}}
where {\small \texttt{\color{greencode}[domain]}} is replaced with the dataset domain (\eg, flower, product, pet, car, \etc). We show \textit{image-description} examples in ~\cref{fig:text-examples}.\\
\noindent\textbf{Group-description} ($\batchcaption$) captures shared visual characteristics between similar images, to combat known limitation of LMM which may miss or hallucinate visual characteristics~\cite{liu2023mllms}.
To generate $\batchcaption$ from image $x$, we randomly sample multiple images with the same pseudo-label as $x$. These are collaged into a single image $x^{\rm group}$ fed to LLAVA which is prompted with:
{\small \texttt{\color{greencode}``Describe the common visual attributes of the [domain] in all the photos concisely, in fewer than 20 words."}}. Examples of such group-descriptions are illustrated in~\cref{fig:text-examples}.

To generate these descriptions, we use LLAVA 1.6~\cite{LLAVA} with a 4-bit quantized Mistral 7B model. This model requires approximately 5GB of GPU memory and takes around 1.2 seconds to generate a single description per image on a Tesla V100 GPU. This makes description generation relatively cost-effective, as we can run five instances of this model in parallel on a Tesla V100 32GB GPU, taking approximately 3.4 hours to generate descriptions for 50k images.

\subsection{Prototype-based CLIP fine-tuning}
\label{sec:proto-learning}

\begin{figure}
    \centering
    \includegraphics[width=\linewidth]{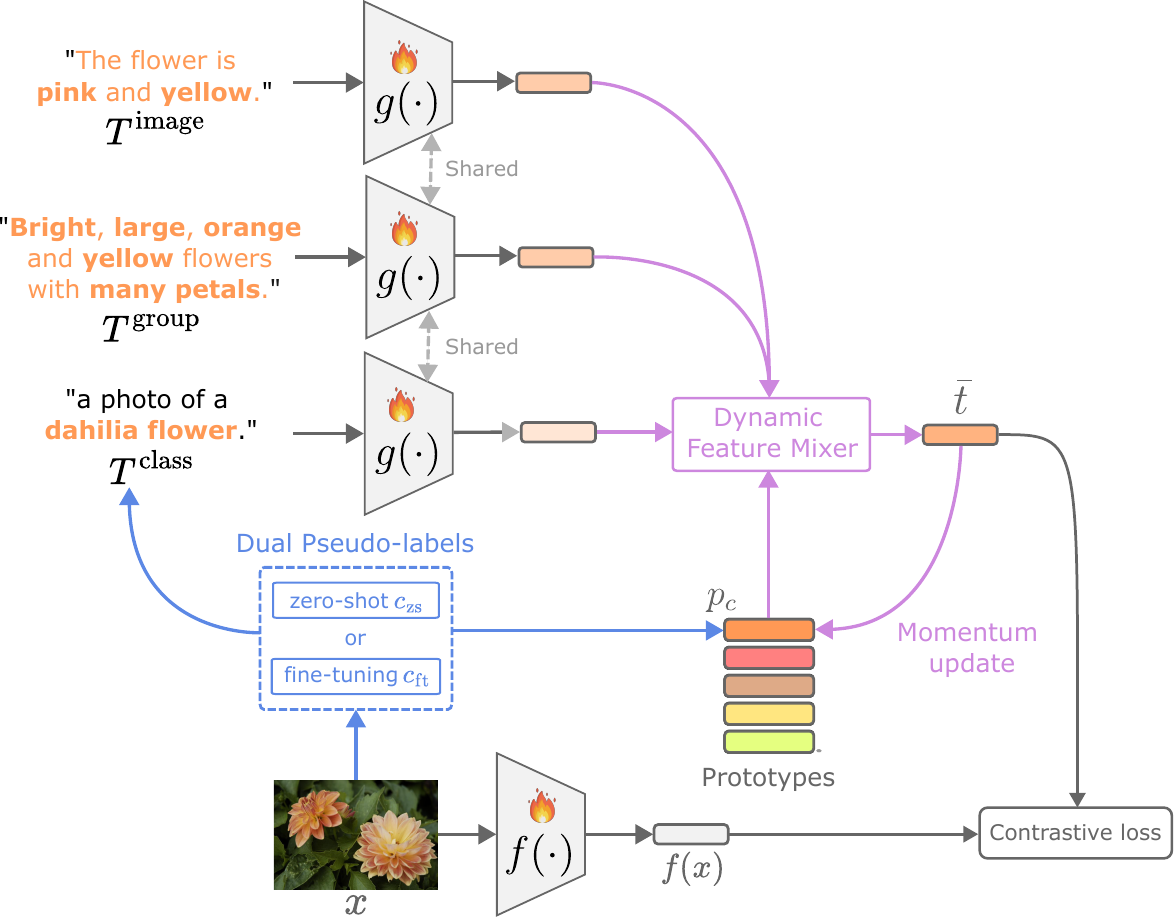}
    \vspace{-0.6em}
    \caption{\textbf{Training.} For image $x$, we predict pseudo-label $c\in\{c_{\rm zs}, c_{\rm ft}\}$ and create three type of descriptions per pseudo-label as described in~\cref{sec:text_gen}. Our Dynamic Feature Mixer combines these descriptions with the corresponding prototype $p_c$ to produce a prototype-text embedding $\bar{t}$, which updates the prototype $p_c$. Lastly, the contrastive loss~\cref{eq:contrastive-loss} is computed between $\bar{t}$ and and the image embedding $f(x)$.
    }
    \label{fig:training}
\end{figure}

Adopting directly the generated texts from \cref{sec:text_gen} is ineffective, because the text encoder overfits to the distribution of generated texts, which are noisy by construction due to hallucinations of the LMM and missing details.
We confirm this experimentally in~\cref{tab:text-abl}, rows 4, 5, 6. 
Therefore, we propose a prototype learning approach that is capable of determining the important synthetic texts and learning better class representations from them.
Our approach mixes three key ingredients as shown in~\cref{fig:training}: 
(1) a simple strategy to preserve robustness by leveraging pseudo-labels from both frozen \textit{and} fine-tuning CLIP models; 
(2) a feature mixer that dynamically balances the importance of each text $\templatecaption$, $\imagecaption$ and $\batchcaption$;
(3) a module that updates the prototypes during training, stabilizing the learning process.

\condenseparagraph{Dual Pseudo-labels.}
As in WISE-FT~\cite{wiseft}, we observe that training only with pseudo-labels from the fine-tuning model improves accuracy but at the cost of overfitting to the training distribution. Hence, to preserve robustness, for each image we employ two pseudo-labels $\{c_{\rm zs}, c_{\rm ft}\}$
originating from \textit{both} the zero-shot model ($c_{\rm zs}$) and the fine-tuning model ($c_{\rm ft}$). We later show that this simple strategy offers greater generalization and accuracy.

\condenseparagraph{Prototype Learning.} 
From the generated texts and pseudo-labels, we aim to learn a set of prototypes corresponding to all classes, denoted as $\{p_c\}_{c=1}^{C}$. These prototypes are designed to capture class-specific details of the synthesized texts and pseudo-labels within the CLIP embedding space. 
{First, the prototypes are initialized with features derived from the $\templatecaption$, generated based on its associated class name.}
Then, for an image $x$, we use our dual pseudo-labels from zero-shot and fine-tuning $\{c_{\rm zs}, c_{\rm ft}\}$ to generate two \emph{class-description} texts $\{\templatecaption_{\rm zs}, \templatecaption_{\rm ft}\}$ and select the corresponding prototypes $p_{\rm zs}$ and $p_{\rm ft}$. 
Our feature mixer strategy, detailed below, then combines the two \emph{class-descriptions} with the \emph{image-description} and the \emph{group-description}, therefore obtaining two prototype-text embeddings $\bar{t}_{\rm zs}$ and $\bar{t}_{\rm ft}$, see~\cref{fig:training}. 
We then apply a momentum update to the corresponding prototypes. Finally, we apply two contrastive losses~\cite{FLYP} between the image embedding $f(x)$ and each of the prototype-text embeddings $\bar{t}_{\rm zs}$ and $\bar{t}_{\rm ft}$.

\begin{figure}
	\centering
	\includegraphics[width=.8\linewidth]{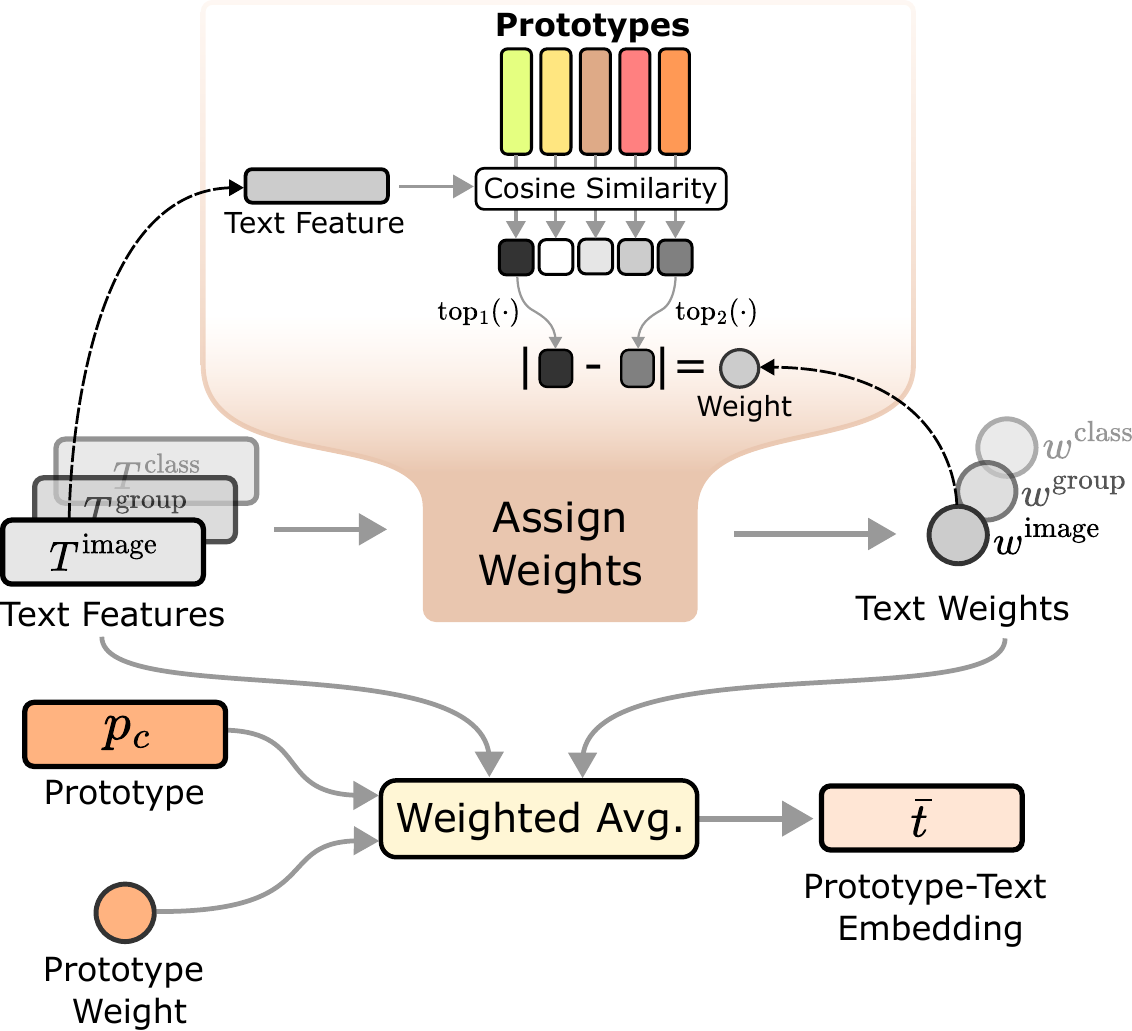} 
 \vspace{-0.6em}
	\caption{\textbf{Dynamic Feature Mixer.} We compute cosine similarities between each text feature and all prototypes. Weights are determined by the difference between the top two similarity scores. We calculate a weighted average of the features and combine it with the prototype (\cref{sec:proto-learning}), creating a representation relevant to the input prototype yet distinct from others.}
	\label{fig:feature_mixing}
\end{figure}

\condenseparagraph{Dynamic Feature Mixer.} 
To compensate for noisy text descriptions, we propose a mechanism that dynamically re-weights the three descriptions as a function of the cosine similarity between each description embedding and corresponding prototype, see~\cref{fig:feature_mixing}.
Intuitively, our goal is to assign higher weights to descriptions uniquely describing a class and lower weights to generic descriptions. 
In the general case, for a text $T$ we first compute the cosine similarities between its CLIP embedding $g(T)$ and each of the prototypes, and obtain its weight $w$ from the difference between the two closest similarities. This writes:
\begin{equation}
w{=}{\rm top_1} \left(\frac{g(T) \cdot p_c}{\|g(T)\| \|p_c\|} \right)_{c=1}^{C} - {\rm top_2} \left( \frac{g(T) \cdot p_c}{\|g(T)\| \|p_c\|} \right)_{c=1}^{C}
\label{eq:weight-compute}
\end{equation}
where ${\rm top_1}(\cdot)$ and ${\rm top_2}(\cdot)$ return the largest and second largest values of the input set, respectively. 
A high weight indicates a large gap between ${\rm top_1}(\cdot)$ and ${\rm top_2}(\cdot)$ similarities, ensuring the text feature is uniquely similar to a single prototype while dissimilar from the rest, as ${\rm top_2}(\cdot)$ value serving as an upper bound for the similarity of the remaining prototypes. Alternatively, we could use the mean or median, but this might result in a text being very similar to a few prototypes while remaining dissimilar to others.
Subsequently, given the set of texts $\{\imagecaption, \batchcaption, \templatecaption \}$ and the weights~$\{w^{\rm image}, w^{\rm group}, w^{\rm class}\}$ computed using~\cref{eq:weight-compute}.
The resulting prototype embedding $\bar{t}$ is defined as
\begin{equation}
\bar{t} = (1-\alpha) \frac{\sum_{i \in I} w^{i} \cdot g(T^i)}{\sum_{i \in I} w^{i}}  + \alpha p_c
\end{equation}
where $I{=}\{{\rm image}, {\rm group}, {\rm class}\}$ and $\alpha$ is the prototype weight. 
We empirically set $\alpha$ to $0.99$ in all experiments to stabilize training, as the prototypes are more reliable than the synthetic text embeddings. Thus, this act a strong regularization mechanism against the noise induced by the synthetic texts. Yet, $\bar{t}$ remains tailored for each image as $\imagecaption$ and $\batchcaption$ differ.
With two pseudo-labels per image, this results in two prototype-text embeddings $\{\bar{t}_{\rm zs}, \bar{t}_{\rm ft}\}$.

\condenseparagraph{Training.} Given an image \( x \), we train both the image encoder \( f(\cdot) \) and text encoder \( g(\cdot) \) using contrastive loss to align the image embedding \( f(x) \) with both the prototype-text embeddings \(\bar{t}_{\text{zs}}\) and \(\bar{t}_{\text{ft}}\), resulting in two losses $\mathcal{L}_{\text{\rm zs}}=\mathcal{L}_{\text{con}}(x, \bar{t}_{\text{zs}})$ and $\mathcal{L}_{\text{\rm ft}}=\mathcal{L}_{\text{con}}(x, \bar{t}_{\text{ft}})$ respectively
with $\mathcal{L}_{\text{con}}(\cdot, \cdot)$ defined as:
\begin{equation}
\begin{split}
\mathcal{L}_{\text{con}}(x, \bar{t}) = & -\frac{1}{N}\sum_{i=1}^N\log\frac{\exp(f(x) \cdot \bar{t}_i / \tau)}{\sum_{j=1}^N \exp(f(x) \cdot \bar{t}_j / \tau)} \\
& -\frac{1}{N}\sum_{i=1}^N\log\frac{\exp(\bar{t}_i \cdot f(x) / \tau)}{\sum_{j=1}^N \exp(\bar{t}_j \cdot f(x_j) / \tau)},
\end{split}
\label{eq:contrastive-loss}
\end{equation}
where $N$ is the batch size and \( \tau \) is the temperature parameter as in \cite{CLIP}. {The first term of~\cref{eq:contrastive-loss} normalizes over text embeddings to match the correct text to an image, while the second normalizes over image embeddings to match the correct image to a text.}
The final loss is $\mathcal{L}_{\text{\rm zs}} + \mathcal{L}_{\text{\rm ft}}$.

\begin{table*}[t]
	\centering
\resizebox{.95\textwidth}{!}{%
	\setlength{\tabcolsep}{0.002\linewidth}
    \begin{tabular}{ l|c|cccccccccc }
        \toprule
        Method & Average & EuroSAT & Sun397 & Food101 & Flower102 & DTD & FGVC & Oxford Pets & Cars & UCF101 & Caltech101 \\
        \midrule
        {Oracle} & {81.76} & {94.46} & {77.45} & {85.01} & {87.90} & {76.65} & {37.95} & {92.42} & {90.21} & {79.49} & {96.02} \\
        \midrule
{LLAVA zero-shot} & {27.23} & {44.78} & {15.74} & {29.81} & {6.58} & {20.27} & {3.18} & {28.92} & {3.38} & {44.25} & {75.38} \\
Pre-trained CLIP &  67.49 & 42.95 & 68.20 & \second{78.65} & \second{71.30} & 55.32 & \second{23.79} & 87.30 & \best{88.25} & 64.37 & \second{94.73}  \\
ReCLIP~\cite{hu2024reclip} & 68.78 & 49.25 & 69.07 & 77.91 & 71.13 & \second{56.91} & \best{25.92} & 88.50 & \second{87.84} & 68.86 & 92.37  \\
FLYP~\cite{FLYP} + pseudo-label~\cite{pseudolabel} & \second{70.01} & \second{67.12} & \second{70.19} & 76.83 & 68.78 & \best{61.82} & 17.40 & \second{88.96} & 84.19 & \second{69.44} & 94.69 \\
\modelname (ours) & \best{72.23} & \best{80.27} & \best{70.68} & \best{79.63} & \best{71.94} &	56.26 &	22.02 &	\best{89.21} &	87.40 &	\best{70.08} &	\best{94.77} \\
        \bottomrule
    \end{tabular}}
    \vspace{-0.6em}
    \caption{Top-1 accuracy on 10 classification datasets. We report the results for {five baselines} and our method. {The `Average' column shows the average results across all datasets.} \best{Best}/\second{$2^{\rm nd}$ best}.}
    \label{tab:main-results}
\end{table*}

\condenseparagraph{Momentum update prototypes.} For a pseudo-label $c$, we derive the corresponding prototype-text embedding $\bar{t}$ for each image. During training, the average prototype-text embedding $\bar{t}_{\rm batch}$ is computed over the images in the batch. Using the pseudo-label $c$, we update the respective prototype $p_c$ with a momentum $\mu$, obtaining the updated embedding $\bar{p}_c = (1-\mu) \bar{t}_{\rm batch} + \mu p_c$, which is then stored back in the prototype bank as the prototype for class $c$. Momentum update works effectively when $\mu \in \{0.99, 0.999, 0.9999\}$~\cite{he2020momentum}. As we fine-tune on smaller dataset with fewer iterations, we set $\mu$ to 0.99 for faster updates of the prototypes. Intuitively, the prototype can be viewed as the running average of the text-prototype embeddings assigned to the class. This process is repeat for each pseudo-label in $\{c_{\rm zs}, c_{\rm ft}\}$.

 \condenseparagraph{Inference.} The predictions for an image $x$ are made by comparing the image embedding $f(x)$, where $f(\cdot)$ is the fine-tuned CLIP image encoder, with $\{p_c\}_{c=1}^{C}$ and taking the prototype with the highest cosine similarity as output.

\section{Experiments}
We evaluate \modelname on the task of fine-tuning on 10 specialized classification datasets, without using any ground truth labels. We use the training set for unsupervised training and use the test set to compute the top-1 accuracy. 

\condenseparagraph{Datasets.} We employ a mixture of datasets covering various specialized domains, including satellite imagery, food dishes, airplane models, and others: EuroSAT~\cite{eurosat}, SUN397~\cite{sun397}, Food101~\cite{food101}, DTD~\cite{dtd}, FGVC~\cite{fgvc}, Oxford Pets~\cite{pets}, Cars~\cite{cars}, UCF101~\cite{ucf101}, Caltech101~\cite{caltech101}, Flower102~\cite{flower102}. These datasets feature specific classes, such as the car model, making the unsupervised fine-tuning setup challenging. We follow the standard train/val/test splits in~\cite{zhou2022learning}. We train $\modelname$ using the combined train and val sets and report its performance on the test set.

\condenseparagraph{Baselines.} We compare our method to {four unsupervised baselines and one fully supervised baseline, which serves as an oracle}. First, we perform zero-shot classification with a pre-trained CLIP model. As in CLIP~\cite{CLIP}, we compute text embeddings for all classes with template {\small \texttt{\color{greencode}``a photo of a [class].''}}. For classification, we compute the cosine similarity between each image and all class text embeddings. Our second baseline, ReCLIP~\cite{hu2024reclip}, also performs fine-tuning without labels but utilizes improved pseudo labels and self-training. 
However, ReCLIP primarily focuses on experiments conducted in a \emph{transductive} manner, which involves training and evaluating on the test split of each dataset. 
To ensure a fair comparison, we retrained ReCLIP using the same CLIP-based model and identical dataset splits as our method.
Third, we combined FLYP~\cite{FLYP} with pseudo-labeling~\cite{pseudolabel} for unsupervised fine-tuning, as the original method relies on supervised fine-tuning. Note that we use FLYP without weight ensembling to maintain a fair comparison with ReCLIP, which also does not employ weight ensembling. 
{Finally, we add "LLAVA zero-shot" baseline which prompts LLAVA to classify the image from a given list of classes, using the following prompt}
{\small \texttt{\color{greencode}``Select the most appropriate category for the image from the following options:[options]. Write only the category name."}}
{, where options is replaced with the list of class names.}
{For the supervised baseline, we train FLYP using ground-truth labels, serving as an oracle.}
The evaluation is performed in a zero-shot fashion, like zero-shot CLIP. Since our method uses prototypes, no class template embeddings have to be computed, and we directly use the prototype vectors. For all baselines and ours, we use OpenCLIP~\cite{openclip}, the open-source implementation of CLIP~\cite{CLIP}, with a ViT/B-32 architecture, pre-trained on the LAION-2B dataset. Performance is reported based on the last epoch since we have no supervision signal. Additional implementation details are in~\cref{sec:supp:implementation}.

\subsection{Results}

\begin{table*}
	\centering
\resizebox{.95\textwidth}{!}{
\setlength{\tabcolsep}{0.004\linewidth}
    \begin{tabular}{ l|ccc|c|cccccccccc }
        \toprule
         & $\templatecaption$ & $\imagecaption$ & $\batchcaption$ & Average & EuroSAT & Sun397 & Food101 & Flower102 & DTD & FGVC & Oxford Pets & Cars & UCF101 & Caltech101  \\
        \midrule
        1 & \checkmark & \checkmark & \checkmark &  \best{72.23} & \best{80.27} & \best{70.68} & 79.63 & 71.94 &	56.26 &	\best{22.02} &	89.21 &	\second{87.40} &	\best{70.08} &	94.77  \\
        2 & \checkmark & \checkmark &  & 70.74 &  \second{79.98} & 64.85 & 75.52 & \best{72.31} & \second{57.03} & 16.44 & 89.45 & 87.09 & 69.50 & \best{95.21}  \\
        3 & \checkmark & & & 70.67 & 78.22 & 59.79 & \best{81.52} & 71.21 & 56.74 & 16.50 & \best{90.00} & \best{87.89} & \second{69.84} & \second{94.97}  \\
        4 & & \checkmark & \checkmark & 55.97 & 64.75 & 63.28 & 76.73 & 50.95 & 48.76 & 9.00 & 64.51 & 32.98 & 60.45 & 88.24  \\
        5 &  & \checkmark & & 52.37 & 44.31 & 62.54 & 77.00 & 48.44 & 43.09 & 7.89 & 56.75 & 33.86 & 58.58 & 91.24  \\
        6 &  & & \checkmark & 53.52 & 59.35 & 65.00 & 77.06 & 31.67 & 49.05 & 9.57 & 64.49 & 25.99 & 66.90 & 86.09  \\
        {7} & {\checkmark} & & {\checkmark} & {\second{71.63}} & {79.68} & {\second{70.07}} & {\second{80.16}} & {\second{71.99}} & {\best{57.47}} & {\second{18.00}} & {\second{89.81}} & {86.15} & {68.84} & {94.12} \\

        \bottomrule
    \end{tabular}}
    \vspace{-0.6em}
    \caption{\textbf{Impact of generated texts.} Best performance is achieved when using all types of descriptions. \best{Best}/\second{$2^{\rm nd}$ best}.}
    \label{tab:text-abl}
\end{table*}

The main results with top-1 accuracy on the 10 datasets are shown in~\cref{tab:main-results}. Across all datasets, \modelname improves the average top-1 accuracy of CLIP by 4.74 points. {Furthermore, it outperforms all unsupervised baselines}, including the recently published ReCLIP~\cite{hu2024reclip} and our proposed baseline that integrates FLYP~\cite{FLYP} with pseudo-labeling~\cite{pseudolabel}, by 3.45 and 2.22 points, respectively. 
Interestingly, FLYP + pseudo-label outperforms ReCLIP, likely due to the robustness and effectiveness of fine-tuning both image and text encoders with contrastive loss, instead of just the image encoder with cross-entropy loss, as demonstrated in FLYP~\cite{FLYP}.
{Notably, LLAVA zero-shot has low overall performance, which could be attributed to LLAVA being trained in generative autogressive manner, thus not optimal for discriminative tasks.}
{Lastly, the oracle is shown on the first line by training FLYP with ground-truth labels. The 9.53-point average performance gap between the fully supervised oracle and unsupervised \modelname highlights room for improvement. Still, \modelname performs competitively, narrowing the gap across multiple datasets, particularly on Oxford Pets, Cars, and Caltech101, to less than 3\%.}

\begin{figure}
    \begin{minipage}[t]{0.48\linewidth}
        {\centering Class: Forest (Eurosat) \par}
        \vspace{0.2em}
         \begin{tikzpicture}
            \node[anchor=south west,inner sep=0] (image) at (0,0) {\includegraphics[width=1.0\textwidth]{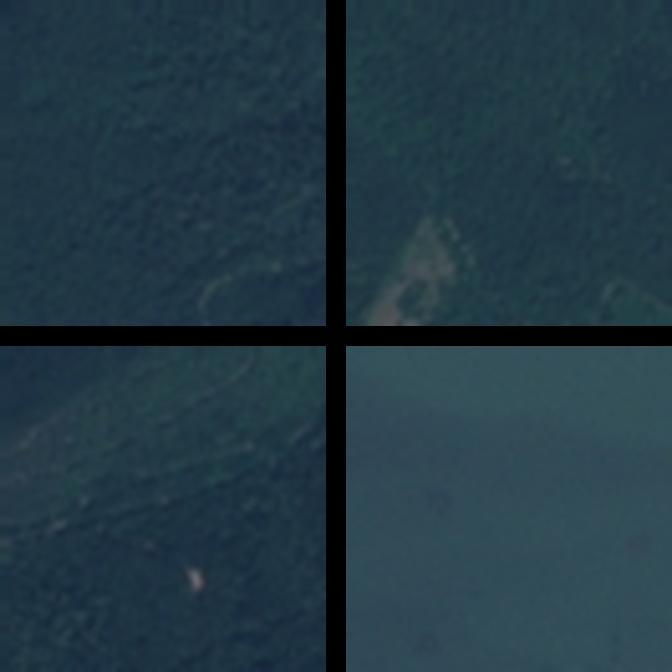}};
            \begin{scope}[x={(image.south east)},y={(image.north west)}]
                \draw[red,thick] (0.01,0.01) rectangle (0.49,0.49);
            \end{scope}
        \end{tikzpicture}
        $\batchcaption$: Dark blue, green, and brown colors, indicating water, vegetation, and land.
        
        $\imagecaption$: The image shows a large body of water with no visible land use.
    \end{minipage}
    \hfill
    \begin{minipage}[t]{0.48\linewidth}
        {\centering Class: Banded (DTD) \par}
        {\vspace{0.2em}}
         \begin{tikzpicture}
            \node[anchor=south west,inner sep=0] (image) at (0,0) {\includegraphics[width=1.0\textwidth]{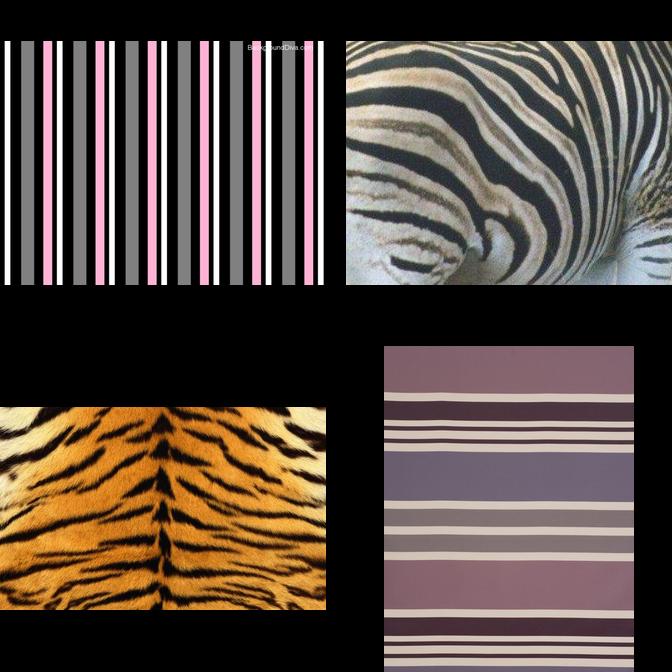}};
            \begin{scope}[x={(image.south east)},y={(image.north west)}]
                \draw[red,thick] (0.49,0.01) rectangle (1.0,0.49);
            \end{scope}
        \end{tikzpicture}
        $\batchcaption$: Stripes, zebra, animal print, geometric shapes, lines, and bold colors.
        
        $\imagecaption$: Striped pattern.
    \end{minipage}
   
    \vspace{0.5em}
    \begin{minipage}[t]{0.48\linewidth}
        {\centering Class: Indoor Factory (SUN397)\par}
         \begin{tikzpicture}
            \node[anchor=south west,inner sep=0] (image) at (0,0) {\includegraphics[width=1.0\textwidth]{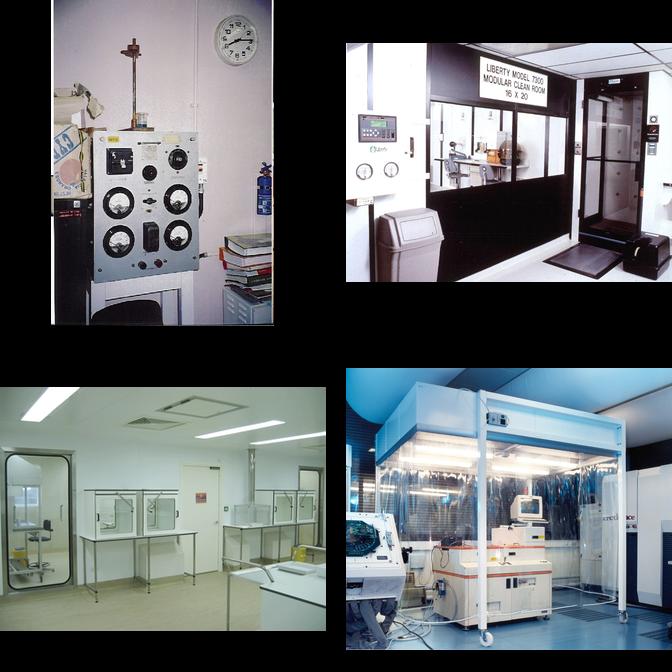}};
            \begin{scope}[x={(image.south east)},y={(image.north west)}]
                \draw[red,thick] (0.51,0.51) rectangle (1.0,1.0);
            \end{scope}
        \end{tikzpicture}
        $\batchcaption$: Industrial factory with white walls, industrial equipment, machinery, clocks, and doors.
        
        $\imagecaption$: Large industrial building with doors.
    \end{minipage}
    \hfill
    \begin{minipage}[t]{0.48\linewidth}
        {\centering Class: Apply Eye Makeup (UCF101) \par} 
         \begin{tikzpicture}
            \node[anchor=south west,inner sep=0] (image) at (0,0) {\includegraphics[width=1.0\textwidth]{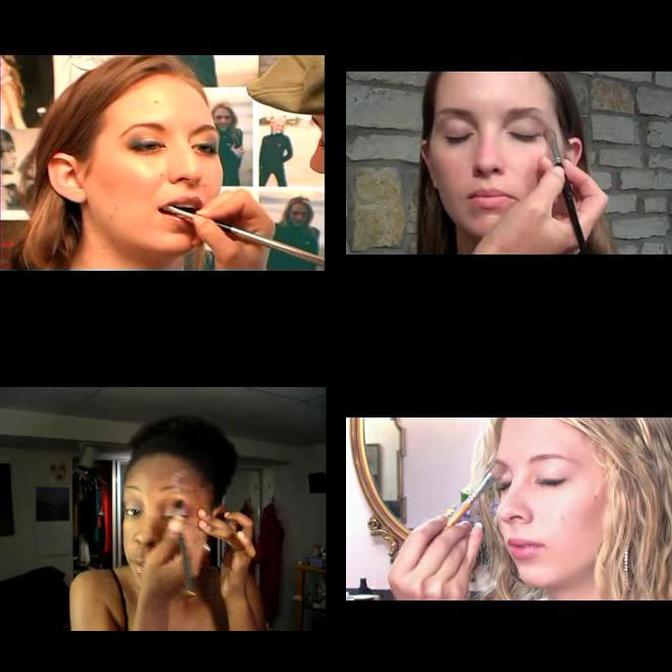}};
            \begin{scope}[x={(image.south east)},y={(image.north west)}]
                \draw[red,thick] (0.01,0.51) rectangle (0.49,1.0);
            \end{scope}
        \end{tikzpicture}
        $\batchcaption$: Makeup application, close-up, hands holding tools.
        
        $\imagecaption$: Girl applying makeup.
    \end{minipage}
    \caption{\textbf{Examples of generated captions.} We either generate a caption from the group of 4 images, by inputting them as tiled single image into LLaVA ($\batchcaption$), or we input a single image to LLaVA ($\imagecaption$). For simplicity, in this figure, we only show a single image caption (highlighted by red bounding box).}
    \label{fig:text-examples}
\end{figure}
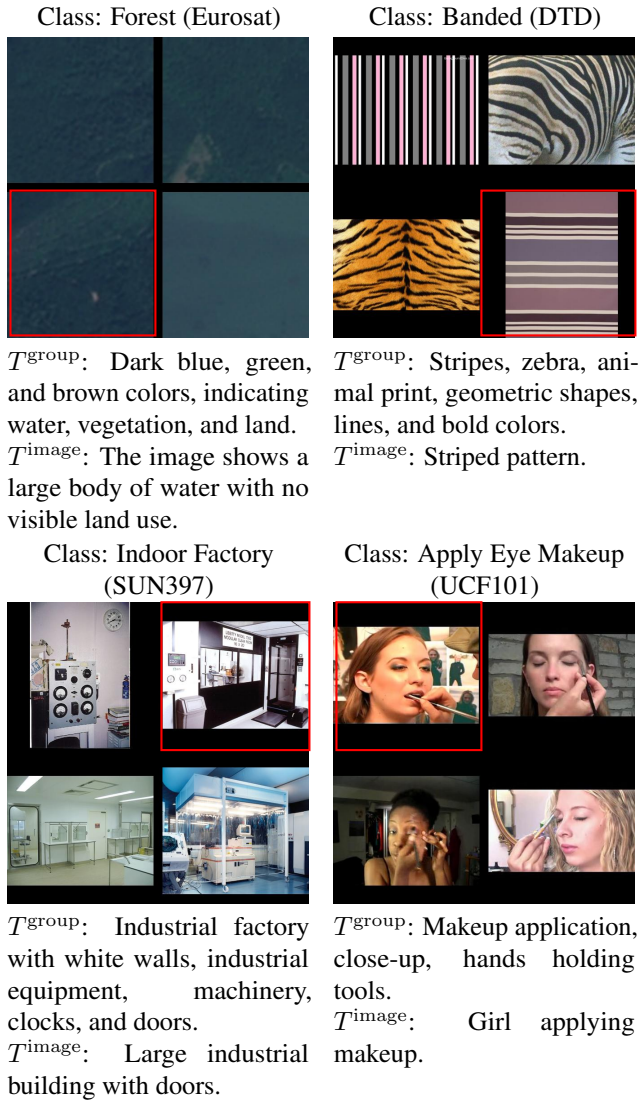

In~\cref{fig:text-examples}, we show examples of generated \emph{image-description} $\imagecaption$ and \emph{group-description} $\batchcaption$. Overall, $\batchcaption$ offers more comprehensive and contextual information. For instance, in the top-right example, $\imagecaption$ is simply "striped pattern," whereas $\batchcaption$ provides richer details, including "zebra, animal print, geometric shape, lines". This trend is evident in other examples as well. For example, in the bottom-right example, $\imagecaption$ is "Girl applying makeup," while $\batchcaption$ elaborates with "Makeup application, close-up, hands holding tools." Additionally, image-description fails to capture "forest" in the top-left example, describing it merely as "The image shows a large body of water with no visible land use." In contrast, $\batchcaption$ includes relevant details such as "vegetation, land, green and brown colors."

\subsection{Ablations} 
\label{sec:ablations}

\begin{table*}
	\centering
\resizebox{.95\textwidth}{!}{
	\setlength{\tabcolsep}{0.003\linewidth}
    \begin{tabular}{ l|c|cccccccccc }
        \toprule
        Method & Average &  EuroSAT & Sun397 & Food101 & Flower102 & DTD & FGVC & Oxford Pets & Cars & UCF101 & Caltech101  \\
        \midrule
        \modelname (ours) & \best{72.23} & \best{80.27} & \best{70.68} & \second{79.63} & \second{71.94} &	56.26 &	22.02 &	\best{89.21} &	\best{87.40} &	\second{70.08} &	\best{94.77} \\
        w/o Dynamic Feature Mixer & 70.23 & 66.04 & 69.41 & \best{80.18} & \best{72.51} & 54.85 &	\second{23.43} &	87.14 &	85.37 &	69.02 &	94.32  \\
        w/o $\mathcal{L}_{\rm ft}$ & 68.26 & 47.06 & \second{69.80} & 79.23 & 70.48 & \second{56.97} & \best{23.82} & \second{87.65} & \second{87.19} & 65.77 & \second{94.60} \\
        w/o $\mathcal{L}_{\rm zs}$ &  \second{70.58} & \second{76.96} & 68.18 & 70.29 & 71.01 & \best{61.05} & 19.89 & 87.49 & 86.13 & \best{70.39} & 94.36 \\
        w/o Momentum Update &  45.72 & 31.19 & 56.17 & 68.08 & 57.41 & 31.32 & 13.41 & 13.03 & 43.31 & 54.69 & 88.56 \\
        \bottomrule
    \end{tabular}}
    \vspace{-0.6em}
    \caption{\textbf{Method ablation.} All components contribute to the best performance. \best{Best}/\second{$2^{\rm nd}$ best}.}
    \label{tab:method-ablation}
\end{table*}

\condenseparagraph{Different types of synthetic descriptions.}
\cref{tab:text-abl} illustrates the impact of different generated texts on overall performance. We observe that all texts are essential for achieving the best performance. 
{Specifically, excluding the \emph{image-description} reduces the average performance across all datasets by 0.6 (row 1 vs. row 7). The impact of removing the \emph{group-description} is even more significant with a 1.49 points reduction (row 1 vs. row 2).}
Additionally, omitting both the \emph{image-description} and \emph{group-description} results in an even larger loss of 1.56 points (comparing row 1 to row 3). Rows 4, 5, and 6 show that relying solely on synthetic texts causes a drop in performance due to the noise and inaccuracies introduced by the generated descriptions.

\condenseparagraph{Dynamic Feature Mixer.} We ablate our Dynamic Feature Mixer in~\cref{tab:method-ablation} (row "w/o Dynamic Feature Mixer") by setting all the text weights to $1.0$, so that all texts contribute equally. The average performance drops by 2 points, with significant decreases on multiple datasets, such as -14.23 on EuroSAT, -1.41 on DTD, and -2.03 on Cars. This demonstrates that our Dynamic Feature Mixer module effectively assigns relevant weights to the meaningful descriptions.

\condenseparagraph{Dual Pseudo-Labels.} Best performance is achieved using both zero-shot and fine-tuning pseudo-labels $\{c_{\rm zs}, c_{\rm ft}\}$. This is assessed in~\cref{tab:method-ablation} by removing the corresponding losses. 
Removing the zero-shot pseudo-label (row~"w/o $\mathcal{L}_{\rm zs}$") leads to a significant drop across multiple datasets: -3.31 on EuroSAT, -2.5 on SUN397, -9.34 on Food101, and an average decline of -1.65 across all datasets. 
Furthermore, removing the fine-tuned pseudo-labels (row~"w/o $\mathcal{L}_{\rm ft}$") results in an even more substantial average performance drop of -3.97, with particularly notable decreases of -33.21 on EuroSAT and -4.31 on UCF101. We conjecture that this is because the zero-shot pseudo-label is more robust, while the fine-tuned pseudo-label has higher accuracy on the training dataset.

\condenseparagraph{Momentum Update.} We ablate the impact of the momentum update by setting $\mu{=}0$, as shown in row "w/o Momentum Update" in~\cref{tab:method-ablation}, therefore directly replacing the prototype by the new weighted text features. 
Without momentum update, performance declines dramatically, with an average decrease of $-26.51$ across all datasets. Significant declines are observed in many datasets, such as -44.09 on Cars, -14.51 on SUN397, and -14.53 on Flower102. This substantial drop is attributed to the high variance in the prototypes due to the noisy generated texts.

\condenseparagraph{Incorrect images in generating $T^{\rm group}$.} We analyze how incorrect images within a group affect the generated group-description $T^{\rm group}$. We test groups of 4 images with different number of correct images, selected using ground-truth labels.  For 1, 2, 3 and 4 correct images, this results in top-1 accuracy of 72.48, 72.61, 72.72, and 72.64, respectively, averaged across all datasets. Performance improves slightly with more correct images. Notably, our method using pseudo-labels achieves a performance of 72.23, which is competitive with the ground-truth label selection. This demonstrates tolerance to noise and pseudo-label inaccuracies within the image group. Detailed performance is provided in~\cref{tab:supp:text-abl-num-correct}.

\condenseparagraph{Number of images per group.} We analyze the impact of increasing the number of images per group on performance by testing groups with 2, 4, 8, and 16 images, resulting in average performance scores of 71.55, 72.23, 72.31, and 72.49, respectively, across all datasets. The performance improves with more images per group, likely because the probability of including the correct images increases. The most notable improvement occurs when increasing the number of images per group from 2 to 4, with score rising from 71.55 to 72.23. This is because achieving a majority with only 2 images requires 100\% accuracy, whereas larger groups can tolerate some errors while still maintaining a correct majority. Performance plateaus after 4 images, possibly due to the fixed input resolution of LLAVA~\cite{LLAVA}, leading to lower per-image resolution as the number of images increases. We provide the performance for all datasets in~\cref{tab:supp:text-abl-n-images}.

\begin{figure}
    \centering
    \includegraphics[width=0.35\textwidth]{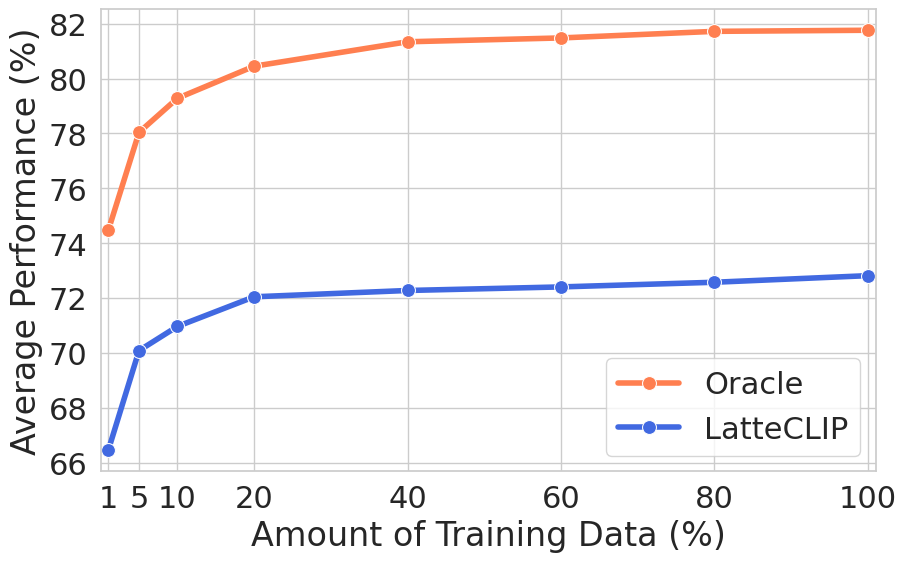}
    \vspace{-0.6em}
    {
    \caption{{\textbf{Impact of amount of training data.} Average top-1 accuracy on 10 datasets while varying the amount of training data.}}
    }
    \label{fig:varying-data}
\end{figure}

\condenseparagraph{{Impact of amount of training data.}}  {\cref{fig:varying-data} illustrates the effect of training data size on the average top-1 accuracy of \modelname and oracle across 10 datasets. Despite being unsupervised, \modelname exhibits strong robustness to varying amounts of training data, comparable to an oracle. Specifically, \modelname's performance drops only 0.77/6.36 on 20\%/1\% data, respectively, compared to 7.28/1.30 of the oracle. Overall, more data improves performance but diminishes notably for both after 20\%.
}

\begin{table*}[h]
	\centering
	\setlength{\tabcolsep}{0.003\linewidth}
    \begin{tabular}{ ll|c|cccccccccc }
        \toprule
        label type & \#correct & Avg. &  EuroSAT & Sun397 & Food101 & Flower102 & DTD & FGVC & Oxford Pets & Cars & UCF101 & Caltech101  \\
        \midrule 
        Pseudo (ours) & N/A & 72.23 & 80.27 & \best{70.68} & \second{79.63} & 71.94 &	56.26 &	\best{22.02} &	89.21 &	87.40 &	\second{70.08} &	94.77 \\
        \midrule
        \multirow{4}{*}{Ground-truth} & 1 &  72.48 & 80.02 & 69.19 & 79.04 & \second{72.88} & \best{61.11} & 20.55 & \best{89.62} & 87.35 & \best{70.16} & 94.89 \\
        & 2 &  72.61 & \best{81.28} & 69.73 & 79.13 & 72.55 & \second{60.82} & 20.76 & \second{89.51} & \second{87.53} & 69.65 & \second{95.13} \\
        & 3 & \best{72.72} & \second{80.81} & 70.28 & \best{79.80} & 72.72 & 60.28 & \second{21.87} & 89.48 & 87.29 & 69.52 & \best{95.17} \\
        & 4 & \second{72.64} & 80.40 & \best{70.54} & 78.79 & \best{72.96} & 60.17 & 21.42 & \best{89.62} & \best{87.96} & 70.00 & 94.56 \\
        \bottomrule
    \end{tabular}
    \vspace{-0.6em}
    \caption{Impact of varying the number of correctly chosen images based on ground-truth labels when using 4 images for \emph{group-description} generation. Our approach yields comparable performance despite relying solely on pseudo-labels for image selection.}
    \label{tab:supp:text-abl-num-correct}
\end{table*}

\begin{table*}[h]
	\centering
	\setlength{\tabcolsep}{0.004\linewidth}
    \begin{tabular}{ l|c|cccccccccc }
        \toprule
        \#Images & Average &  EuroSAT & Sun397 & Food101 & Flower102 & DTD & FGVC & Oxford Pets & Cars & UCF101 & Caltech101  \\
        \midrule 
        2 & 71.55 & \best{80.74} & 69.36 & 76.03 & 71.24 & 56.03 & 21.12 & \second{89.29} &  87.32 & 69.88 & \second{94.52} \\
        4 & 72.23 & 80.27 & \best{70.68} & \best{79.63} & 71.94 & 56.26 & \second{22.02} & 89.21 & 87.40 & \best{70.08} & \best{94.77} \\
        8 & \second{72.31} & 79.90 & 69.90 & \second{79.55} & \second{73.04} & \second{57.69} & 22.00 & 89.18 & \second{87.65} & 69.71 & 94.44 \\
        16 & \best{72.49} & \second{80.67} & \second{70.18} & 78.24 & \best{73.20} & \best{58.64} & \best{22.28} & \best{89.53} & \best{87.85} & \second{70.05} & 94.28 \\
        \bottomrule
    \end{tabular}
    \vspace{-0.6em}
    \caption{Number of images per group impact on generating group-descriptions. Overall, more images improve performance due to richer information and increased robustness against the inclusion of incorrect images. However, the performance plateaus on some datasets, e.g., UCF101 or SUN397, could be attributed to LLAVA's fixed resolution, resulting in lower resolution per image when using more images.}
    \label{tab:supp:text-abl-n-images}
\end{table*}

\section{Conclusion}
$\modelname$ is a novel method for unsupervised CLIP fine-tuning on specialized datasets where human annotations are costly or require expert knowledge. Leveraging LMMs, \modelname generates rich and expressive synthetic textual descriptions at various levels of contextual granularity, including \emph{image-description}, \emph{group-description}, and \emph{class-description}. To effectively learn from these potentially noisy descriptions, we propose a prototype learning framework with three key elements: (1) dual pseudo-labels from frozen and fine-tuning CLIP models; (2) a Dynamic Feature Mixer for optimal text feature weighting; and (3) momentum update to enhance training stability. \modelname surpasses comparable baselines on average across all datasets.

\newpage
\condenseparagraph{Acknowledgment.} 
The research was conducted during Quan's internship at Amazon. The research was also supported by the ANR project SIGHT (ANR-20-CE23-0016) and SAMBA collaborative project co-funded by BpiFrance in the Investissement d’Avenir Program. Computation was performed using HPC resources from GENCI–IDRIS (AD011012808R2, AD011014102R1). We thank Ajanthan Thalaiyasingam and Mohammad Fahes for their insightful suggestions. We also extend our gratitude to Mohammad Fahes and Ivan Lopes for their thorough proofreading.

\appendix
\section*{Appendices}

\section{{Limitations}} 
\label{sec:supp:limitation}
{
Despite promising results, \modelname considers a limited number of description types. Expanding description generation to include more contextual levels, such as scenes, objects, and attributes, would provide richer contextual information. Additionally, our performance is constrained by the underlying LMM model, and improvements could be made with better models in the future. Lastly, it is unclear why the method improves on some datasets but not others. Understanding this discrepancy could lead to better methods.
}

\section{Implementation details}
\label{sec:supp:implementation}
We implement \modelname based on the standard fine-tuning pipeline of OpenCLIP~\cite{openclip} using the VIT-B/32 model. The hyperparameters used are the default ones provided in OpenCLIP~\cite{openclip}, except for batch size and learning rate. We use a batch size of 512 and a learning rate of 1e-7 for the datasets Caltech101~\cite{caltech101}, DTD~\cite{dtd}, Eurosat~\cite{eurosat}, FGVC~\cite{fgvc}, Oxford Pets~\cite{pets}, Cars~\cite{cars}, Flower102~\cite{flower102}, and UCF101~\cite{ucf101}. For the datasets Food101~\cite{food101} and SUN397~\cite{sun397}, we use a learning rate of 1e-6. \modelname is trained for min\{2000 iterations, 50 epochs\}.

For FLYP~\cite{FLYP}, we reimplement it based on its official implementation\footnote{\url{https://github.com/locuslab/FLYP}} and OpenCLIP~\cite{openclip}, as its idea is intuitive and simple: fine-tuning using contrastive loss with class templates instead of cross-entropy loss. We use the same OpenCLIP-based model and training hyperparameters as \modelname. The pseudo-labels are recalculated after every weight update, following~\cite{pseudolabel}. 

For ReCLIP~\cite{hu2024reclip}, we use the official implementation\footnote{\url{https://github.com/michiganleon/ReCLIP_WACV}}, but substitute OpenCLIP as the base CLIP model to ensure a fair comparison across all methods. While ReCLIP is designed for transductive learning (train/test on test set), as shown in the paper and by its official implementation, we adapt it to our experimental setup. Specifically, we retrain and evaluate ReCLIP using identical dataset splits as \modelname.

\begin{figure*}[h]
\setlength{\tabcolsep}{0.005\linewidth}
\newcolumntype{P}[1]{>{\centering\arraybackslash}m{#1}}
\centering
    \begin{tabular}{P{0.10\textwidth} P{0.18\textwidth} | P{0.14\textwidth} P{0.25\textwidth} | P{0.15\textwidth} | P{0.11\textwidth}}
        \toprule
        $x$ & $\imagecaption$ & $x^{\rm group}$ & $\batchcaption$ & pseudo-labels & GT \\
        \midrule
        \includegraphics[width=\linewidth]{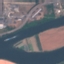} & Buildings and green spaces. & \includegraphics[width=\linewidth]{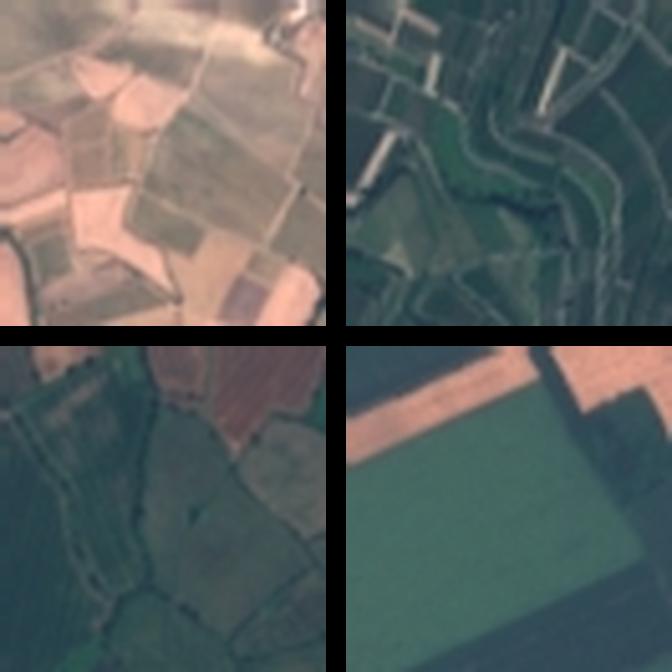} & Green, brown, and blue colors, indicating vegetation, soil, and water. & $c_{\rm zs}$: permanent crop land, $c_{\rm ft}$:  river & river (Eurosat) \\
        \includegraphics[width=\linewidth]{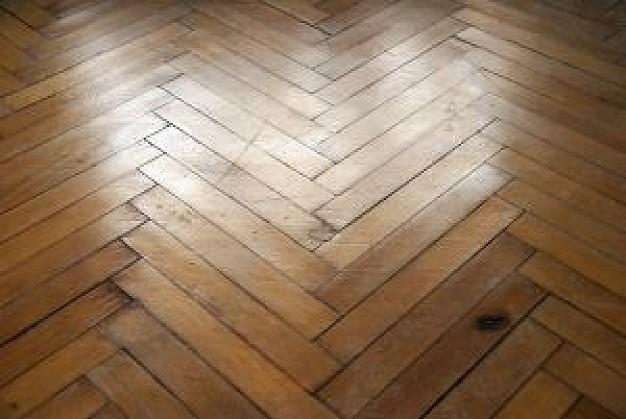} & The texture in the photo is a wooden floor with a herringbone pattern. & \includegraphics[width=\linewidth]{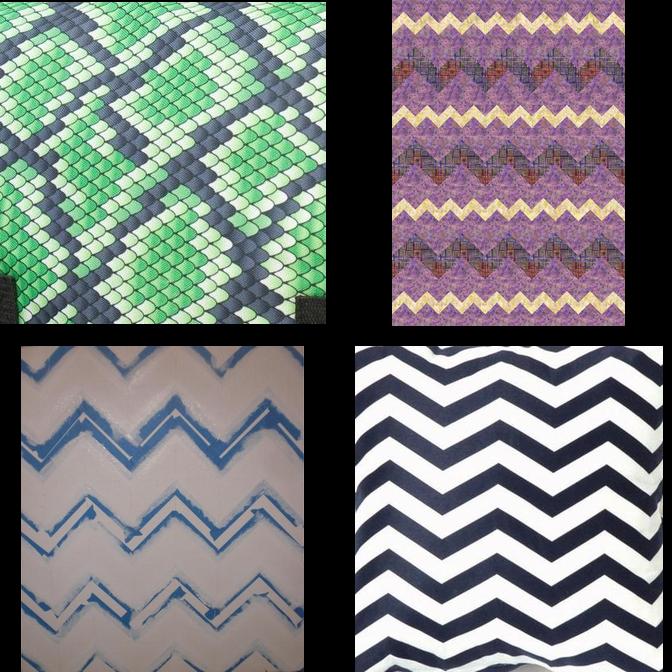} & Zigzag patterns, geometric shapes, and vibrant colors. & $c_{\rm zs}$: zigzagged, $c_{\rm ft}$: grooved & zigzagged (DTD) \\
        \includegraphics[width=\linewidth]{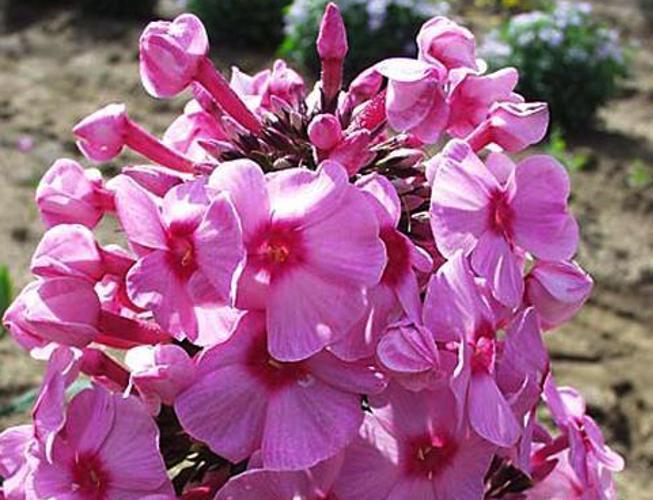} & The pink primrose flower in the photo is a beautiful and vibrant display of nature's beauty. & \includegraphics[width=\linewidth]{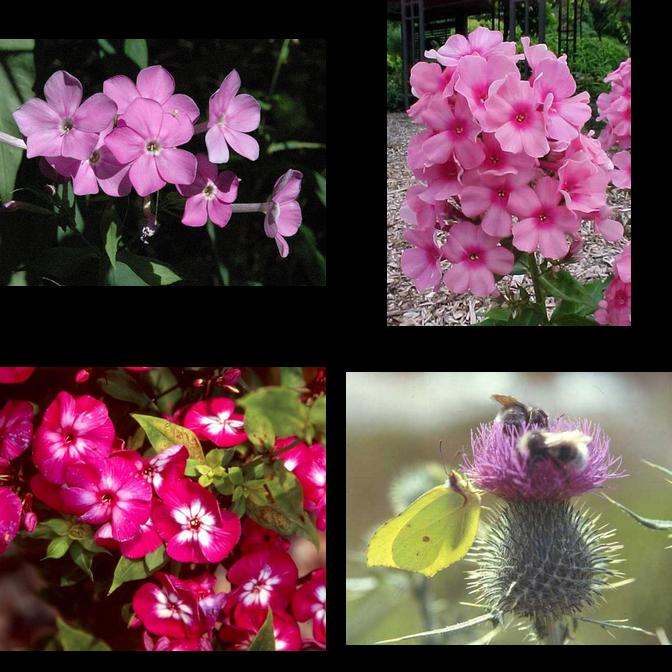} & Purple and yellow petals, green stems, multiple layers of petals. & $c_{\rm zs}$: pink primrose, $c_{\rm ft}$: silverbush & garden phlox (Flower102) \\
        \includegraphics[width=\linewidth]{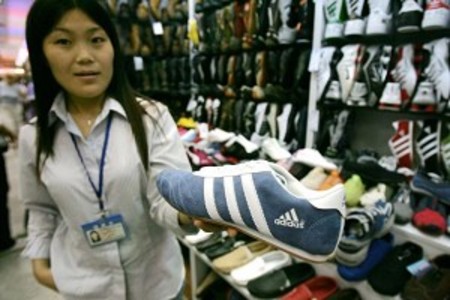} & Woman in white shirt holding blue shoe. & \includegraphics[width=\linewidth]{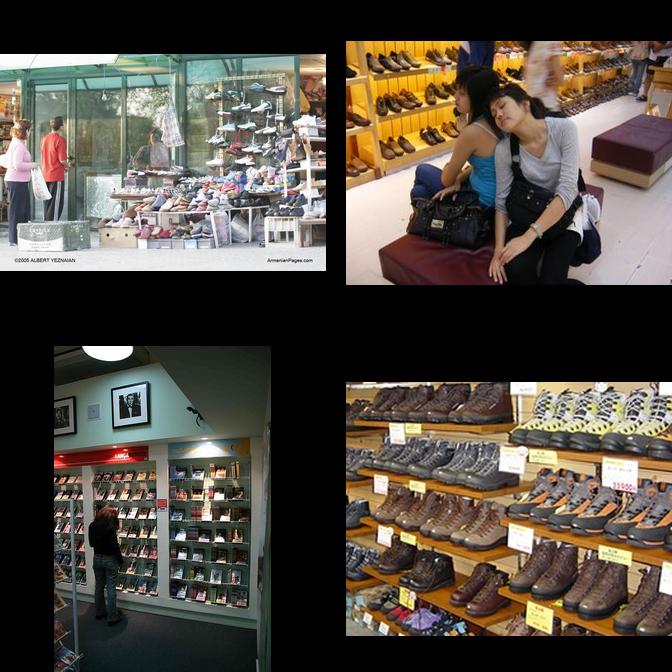} & Shoes, women, shopping, retail, store, display, merchandise, fashion, sales, shopping center, mall, department store, commercial, consumer. & $c_{\rm zs}$: shoe shop, $c_{\rm ft}$: shoe shop & shoe shop (SUN397) \\
        \includegraphics[width=\linewidth]{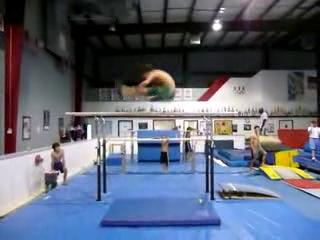} & Person on trampoline. &  \includegraphics[width=\linewidth]{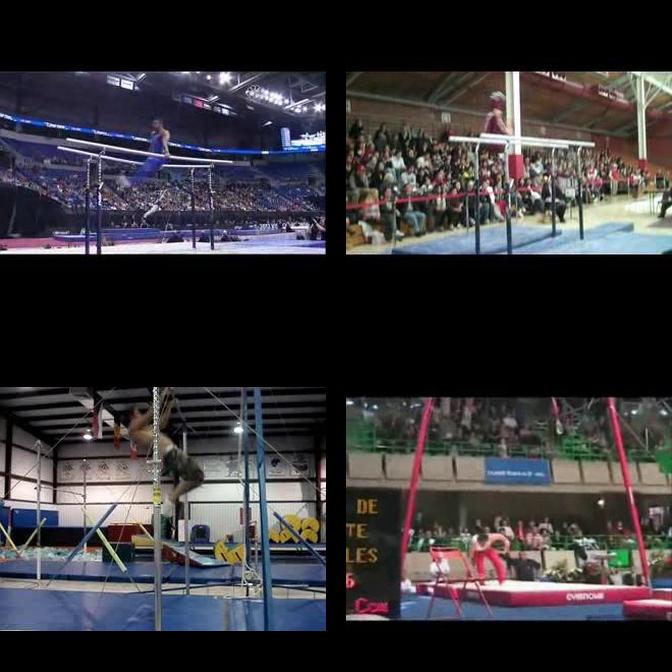} & Gymnastics, acrobatics, high jumps, flips, and aerial stunts. & $c_{\rm zs}$: uneven bars, $c_{\rm ft}$: parallel bars & parallel bars (UCF101) \\
        \bottomrule
    \end{tabular}
    \vspace{-0.6em}
    \caption{Examples of \emph{image-description} $\imagecaption$ generated from image $x$ and \emph{group-description} $\batchcaption$ generated from image group $x^{\rm group}$, and two types of pseudo-labels: zero-shot $c_{\rm zs}$ and fine-tuning $c_{\rm ft}$. The class-description is generated by substituting the pseudo-label $c \in \{c_{\rm zs}, c_{\rm ft}\}$ into a predefined template: \texttt{\color{greencode}``a photo of a [$c$].''}.}
    \label{fig:additional-qual-results}
\end{figure*}

\section{Additional ablations}
\label{sec:supp:ablations}

\condenseparagraph{Incorrect images in generating $T^{\rm group}$.}~\cref{tab:supp:text-abl-num-correct} presents the results across all datasets when varying the number of correct images, which are selected using ground-truth labels, in groups of 4 images used for generating \emph{group-descriptions}. Using more correct images generally leads to improvements in most datasets. However, the average performance gap remains small, demonstrating the robustness of our method to the presence of incorrect images in the group. This robustness is further evidenced by the performance of \modelname, which remains competitive even when relying on pseudo-labels for image selection instead of ground-truth labels.

\condenseparagraph{Number of images per group.}~\cref{tab:supp:text-abl-n-images} analyzes the performance as the number of images per group used for generating \emph{group-description} increases. Generally, more images per group lead to higher performance on most datasets. This is intuitive, as more images provide richer information and a higher chance of including correct images. Using only two images results in the worst performance because selecting a wrong image would significantly impact the outcome, making 50\% or 100\% of the selected images incorrect. Consequently, larger groups are more robust to the inclusion of wrong images. As LLAVA~\cite{LLAVA} has a fixed resolution, adding more images results in lower resolution per image. This could explain the performance plateau on datasets with more image details, such as UCF101 or SUN397.

\section{Additional results}
\label{sec:supp:additional-results}

\condenseparagraph{Examples of LMM-synthetic texts and pseudo-labels.}~\cref{fig:additional-qual-results} illustrates examples of \emph{image-description} $\imagecaption$ and \emph{group-description} $\batchcaption$ generated from individual images $x$ and image groups $x^{\rm group}$, respectively. The figure also presents ground-truth labels (GT) along with pseudo-labels derived from the frozen CLIP model ($c_{\rm zs}$) and the fine-tuning model ($c_{\rm ft}$). Note that the class-description is generated by substituting the pseudo-label $c \in \{c_{\rm zs}, c_{\rm ft}\}$ into a predefined template: \texttt{\color{greencode}``a photo of a [$c$].''}. Combining both types of pseudo-labels increases the chance of capturing the ground-truth label, as each type of pseudo-label is correct for different examples. For instance, $c_{\rm zs}$ is correct for rows 2, 3, and 4, while $c_{\rm ft}$ is correct for rows 1 and 4. Regarding the synthetic description, $T^{\rm group}$ provides richer contextual information, particularly in rows 1, 2, 4, and 5, and contains less hallucinated information compared to $\imagecaption$, as seen in rows 2 and 3, with greater accuracy in rows 1, 4, and 5.

\FloatBarrier  
{\small
\bibliographystyle{ieee_fullname}
\bibliography{main}
}

\end{document}